\documentclass[runningheads]{llncs}

\usepackage[year=2024,ID=14]{eccv}

\usepackage{eccvabbrv}

\usepackage{graphicx}
\usepackage{booktabs}
\usepackage{algorithm}
\usepackage{algpseudocode}
\usepackage{alltt}
\usepackage{multirow}

\usepackage[accsupp]{axessibility}  

\newcommand{\bfD}{\mathbf{D}}

\newcommand{\bfM}{\mathbf{M}}

\newcommand{\bfE}{\mathbf{E}}
\newcommand{\bfL}{\mathbf{L}}
\newcommand{\bfC}{\mathbf{C}}
\newcommand{\bfP}{\mathbf{P}}
\newcommand{\bfI}{\mathbf{I}}

\newcommand{\argmin}{\operatornamewithlimits{argmin}} 

\newcommand{\Exp}[1]{\underset{#1}{\mathbb{E}}}

\usepackage[pagebackref,breaklinks,colorlinks,citecolor=eccvblue]{hyperref}

\usepackage{orcidlink}

\begin{document}

\title{Optimal OnTheFly Feedback Control of Event Sensors} 

\titlerunning{OnTheFly  Control of Event Sensor}

\author{Valery Vishenvskiy\inst{1} \and
Greg Burman\inst{2} \and
Sebastian Kozerke\inst{1}\and 
Diederik Paul Moeys\inst{2} 
}

\authorrunning{V.~Vishnevskiy et al.}

\institute{ETH Zürich, Switzerland \\
\email{\{vishnevskiy, kozerke\}@biomed.ee.ethz.ch}
\and
{Sony Semiconductor Solutions Europe, Sony Europe B.V,\\Stuttgart Laboratory 1, Zürich Office}\\
\email{\{greg.burman, diederik.moeys\}@sony.com}
}

\maketitle

\begin{abstract}
Event-based vision sensors produce an asynchronous stream of events which are triggered when the pixel intensity variation exceeds a predefined threshold.  
Such sensors offer significant advantages, including reduced data redundancy, micro-second temporal resolution, and low power consumption, making them valuable for applications in robotics and computer vision. 
In this work, we consider the problem of video reconstruction from events, and propose an approach for dynamic feedback control of activation thresholds, in which a controller network analyzes the past emitted events and predicts the optimal distribution of activation thresholds for the following time segment.
Additionally, we allow a user-defined target peak-event-rate for which the control network is conditioned and optimized to predict per-column activation thresholds that would eventually produce the best possible video reconstruction.
The proposed \emph{OnTheFly} control scheme is data-driven and trained in an end-to-end fashion using probabilistic relaxation of the discrete event representation.
We demonstrate that our approach outperforms both fixed and randomly-varying threshold schemes by 6-12\% in terms of LPIPS perceptual image dissimilarity metric, and by 49\% in terms of event rate, achieving superior reconstruction quality while enabling a fine-tuned balance between performance accuracy and the event rate.
Additionally, we show that sampling strategies provided by our OnTheFly control are interpretable and reflect the characteristics of the scene.
Our results, derived from a physically-accurate simulator, underline the promise of the proposed methodology in enhancing the utility of event cameras for image reconstruction and other downstream tasks, paving the way for hardware implementation of dynamic feedback EVS control in silicon.  
\end{abstract}

\section{Introduction}

Event-based Vision Sensors (EVS), also known as Dynamic Vision Sensors (DVS)~\cite{lichtsteiner2008a128x128, gallego2019eventbased} or simply event cameras, are bio-inspired imaging sensors which behave similarly to the biological retina. 
Such sensors only detect logarithmic changes in brightness, and transmit these changes continuously and asynchronously as \emph{events}. 
An event is composed of a timestamp, location and polarity, with the last describing whether the change was positive (ON) or negative (OFF). 
The main advantages of such frame-less cameras are (i) a reduction of redundant data, (ii) micro-second temporal resolution, (iii) low power consumption and (iv) High Dynamic Range (HDR), due to the logarithmic compression.

However, the most fundamental and intuitive problem for the EVS workflow is image reconstruction, i.e. recovering a video from the event stream.
Given the reconstructed video, any general-purpose algorithm trained on natural images can be applied to EVS data.
This is however a difficult task due to various EVS non-idealities, such as mismatch across pixels and noise~\cite{lichtsteiner2008a128x128, moeys2018asensitive}, which distort and corrupt the lossy quantization encoding mechanism.
Linear event noise coming from parasitic photocurrents~\cite{nozaki2017temperature}, as well as thermal and shot MOSFET transistor noise mixed with photoreceptor shot noise coming before the logarithmic conversion, make it challenging to process the event stream. For this reason, a naive integration of events and exponentiation such as ~\cite{moeys2017colortemporal} degrades very quickly over time, because a single noise event interpreted as a logarithmic change in brightness has big effects on the reconstructed output.
All these uncertainties in the sensor state pose a great challenge for the model-based reconstruction conducted via solving the \emph{inverse problem}~\cite{zhang2022formulating}.
Such a formulation relies on the temporal event integration, which in practice quickly diverges due the to aforementioned noise sources. 
To overcome this challenge, modern ``event-to-video'' methods employ more sophisticated priors and train recurrent convolutional networks to predict natural scene image from the discretized event stream~\cite{rebecq2019high, Kim2014SimultaneousMA, 6033299, scheerlinck2020fast}. 

\begin{figure}[t]
  \centering
  \begin{minipage}[t]{0.4\linewidth}
   \hspace{0.45\linewidth}{\scriptsize (a)}\\
    \includegraphics[width=\linewidth]{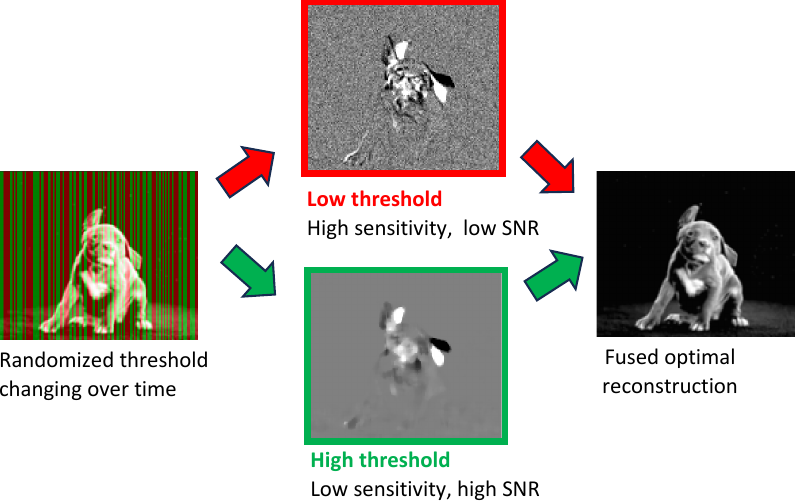}
  \end{minipage}
  \hfill
  \begin{minipage}[t]{0.55\linewidth}
{\hspace{0.45\linewidth} \scriptsize (b) }\\
    \includegraphics[width=\linewidth]{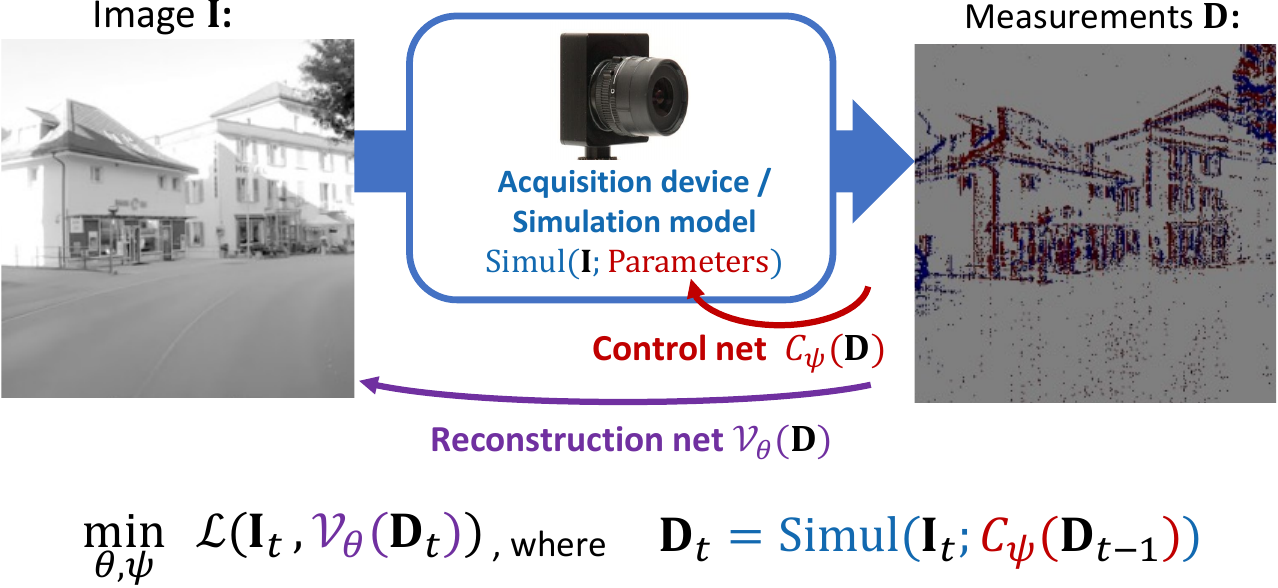}
  \end{minipage}
  \caption{
  (a) Illustration of image reconstruction from events with spatiotemporally-variable parameters providing complimentary incoherence in the data, allowing for an optimal combination.
  (b) Data-driven acquisition optimization, illustrated for EVS.
A ground-truth image $\bfI$ is passed to the simulator ``Simul'', yielding discretized events $\bfD$, which are then mapped back by a reconstruction network $\mathcal{V}_\theta$ to match the initial image.
The control network $\mathcal{C}_\psi$ adjusts the sensor acquisition parameters to achieve the best possible image reconstruction accuracy. 
Reconstruction and control network weights are optimized jointly at the training stage to minimize the target image dissimilarity metric $\mathcal{L}$.
During inference, the simulator can be swapped with a real-camera setup.
  }
  \label{fig:event_fuse_main_idea}
\end{figure}

While deep-learning-based methods achieve remarkable image reconstruction quality, they still operate with the assumption of fixed and predefined EVS sensor parameters. 
However, similar to standard cameras, EVS acquisition can potentially be optimized for specific scenes and scenarios.
For example, high contrast sensitivity can produce too many noisy events in low-light conditions, and different applications might require different trade-offs between the temporal resolution of the data and effective signal-to-noise ratio (SNR).
Finally, exposing EVS to a very dynamic scene, or one with an extremely wide illumination range, can saturate the event transfer circuitry, which would lead to dropping of events and excessive power drain by the sensor, negating most of the advantages attributed to event cameras.
In this case it would be more reasonable to \emph{increase} the activation threshold $\Delta$ to decrease the event rate.
Moreover, assigning different activation thresholds to neighbouring pixels would permit acquiring complementary information with different noise and sensitivity characteristics, which can then be fused together, combining the best properties of each sensing regime (see Fig.~\ref{fig:event_fuse_main_idea}(a)).
In summary, the quantity of information encoded in the data is finite, and limitless refining of a reconstruction algorithm could ultimately result in prior knowledge outweighing the actual data acquired, leading to unfaithful reconstruction and hallucination of image features. 
Therefore, we emphasize that optimizing the way images are encoded is just as crucial as developing efficient reconstruction methods.

In this work, we address the problem of the on-the-fly or optimal \emph{feedback} control of EVS cameras.
Namely, we develop a methodology to dynamically assign per-column activation threshold values $\Delta$ across the EVS pixel array.
The exact values of this control parameter are governed by the sensor control network, which analyzes the events from previous frames and predicts the optimal control for the future frame in order to maximize the amount of encoded data, while keeping the number of generated events as low as possible.
This problem of acquisition optimization can be roughly stated as \emph{forward}-\emph{inverse} model optimization (see Fig.~\ref{fig:event_fuse_main_idea}(b)), and is of interest for many imaging devices, such as standard cameras and medical imaging pipelines.
Nowadays EVS optimization research is at its initial stage~\cite{delbruck2021feedback, graca2023optimal, gracca2023shining} which only deals with a single global control parameter applied to the whole sensor.
In contrast, here we consider a more granular per-column control of the sensor, and tackle the EVS optimization at a higher level, in an end-to-end fashion.
Namely, our control network is task-conditioned: it pursues the best possible sensor control for the given target event rate.
Additionally, our approach allows a synergy between the control and reconstruction network: i.e. the control network is trained not to generate too many events in the areas where the reconstruction network performs sufficiently well.

The main contributions of this work can be summarized as follows:
\begin{itemize}
    \item We develop a model for the self-driven data-dependent event camera parameter control feedback loop.
    \item The method is applied to the problem of image reconstruction, and our proposed control is shown to improve reconstruction accuracy and event bandwidth.
    \item We develop a mechanism to guide the control network over the possible reconstruction accuracy vs. event rate trade-off range, where improved control is achieved for each regime.
    \item The control strategies produced by the network are analyzed and shown to have an intuitive interpretation.
\end{itemize}

We believe that our work can allow for smarter acquisition strategies and further improvement of event camera benefits.
Being capable of always achieving the highest quality in the downstream task is only attainable through sensing as much information as possible, while adapting to the changing environment. 
Our approach is not limited to EVS, but can potentially be generalized to virtually any other sensor whose parameters can be dynamically configured over time.


\section{Related Work}
Compared to the standard cameras, the optimization of EVS control parameters is largely unexplored and is at its initial phase~\cite{delbruck2021feedback}.
Three main EVS control parameters are currently considered for optimization in the literature: (i) activation threshold $\Delta$, which defines the sensitivity of the pixel to contrast-variation for event generation; (ii) refractory period, which defines the \emph{dead time} after an event, i.e. how long the pixel is inactive after event generation; and (iii) photoreceptor bandwidth, which acts as a low-pass filter that effectively smooths the input signal and thus reduces part of the electronic noise.
McReynolds~\emph{et al.}~\cite{mcreynolds2022experimental} experimentally measured effects of these parameters on the output event stream.
Dilmaghani~\emph{et al.}~\cite{dilmaghani2023control} considered various camera use cases and provided qualitative analyses and guidelines on the information presented in the EVS output stream with respect to different sensor configurations, in order to increase sharpness in an integrated event time-bin frame.
The overall conclusion of this research indicates roughly monotonous relation between the event rate and control parameters: e.g. increasing activation threshold or refractory time decreases the output event rate.
Finally, Delbruck~\emph{et al.}~\cite{delbruck2021feedback, graca2023optimal, gracca2023shining} analyzed sensor sensitivity vs. SNR trade-offs from the circuit design perspective. 
They also addressed the problem of basic \emph{dynamic feedback} control of EVS parameters based on the instantaneous event rate: a running average of the event rate is maintained, while the fixed-step controller takes small inverse-proportional steps in the control parameter to achieve the desired event rate. 
This control strategy assumes only a single free control parameter globally for the whole sensor: either activation threshold, refractory period or bandwidth of the photoreceptor.

\textbf{In standard camera} imaging the end-to-end design optimization aims at improving image quality together with downstream tasks.
For example Sitzmann~\emph{et al.}~\cite{sitzmann2018end} consider the problem of optimization of \emph{optical} design of the camera lens.
The lens-dependent blurred image is acquired from the camera and passed to a simple Wiener deconvolution, while the result should match the initial sharp ground truth image. 
This work however assumes that the lens is fixed, so there is no feedback control.
Similarly, it was shown~\cite{chakrabarti2016learning} that optimization of the per-pixel color multiplexing pattern allows improved resolution and color accuracy of the reconstructed image.

Martel~\emph{et al.}~\cite{Martel:2020:NeuralSensors} consider a problem of HDR image reconstruction using a programmable SCAMP-5 vision chip.
They optimize the spatio-temporal shutter function map: i.e. utilize an asynchronous subexposure distribution to extract incoherent but complimentary encoding of neighbouring pixels, hence maximizing the amount of extracted information from the scene. 
However, due to the limitations of SCAMP computational resources, this shutter function map is fixed after training . 
Although the authors argue that different shutter map presets can be applied for different imaging conditions, this mechanism is not elaborated and such an approach would not allow dynamic feedback control.

\textbf{In medical imaging} the design of acquisition methods has undergone extensive research, driven by the enhanced flexibility of acquisition devices and the potential improvements in patient comfort, economic efficiency, and diagnostic quality.
Kellman~\emph{et al.}~\cite{kellman2019physics} enhanced the code-illumination pattern for quantitative phase microscopy. 
They achieved this by unrolling the image reconstruction algorithm and backpropagating through it, which allowed for the subsequent optimization of the light-emitting diodes' intensities, which ultimately improved reconstruction accuracy and speed.

In the area of Magnetic Resonance Imaging (MRI) a substantial amount of work has been conducted to improve the undersampling acquisition pattern for accelerated imaging. 
Different optimization techniques, including greedy search~\cite{sanchez2020scalable}, smooth~\cite{bahadir2019learning} and probabilistic~\cite{vishnevskiylearning, vishnevskiyprobabilistic} relaxation of the binary step function, or continuous frequency sampling trajectories~\cite{wang2022b} have been employed.
Recently, self-supervised approaches that dynamically depend on the acquired data are being developed~\cite{jin2019self, vishnevskiylearning}, potentially leading to optimal patient-specific acquisition strategies.

\begin{figure}
    \includegraphics[width=\linewidth]{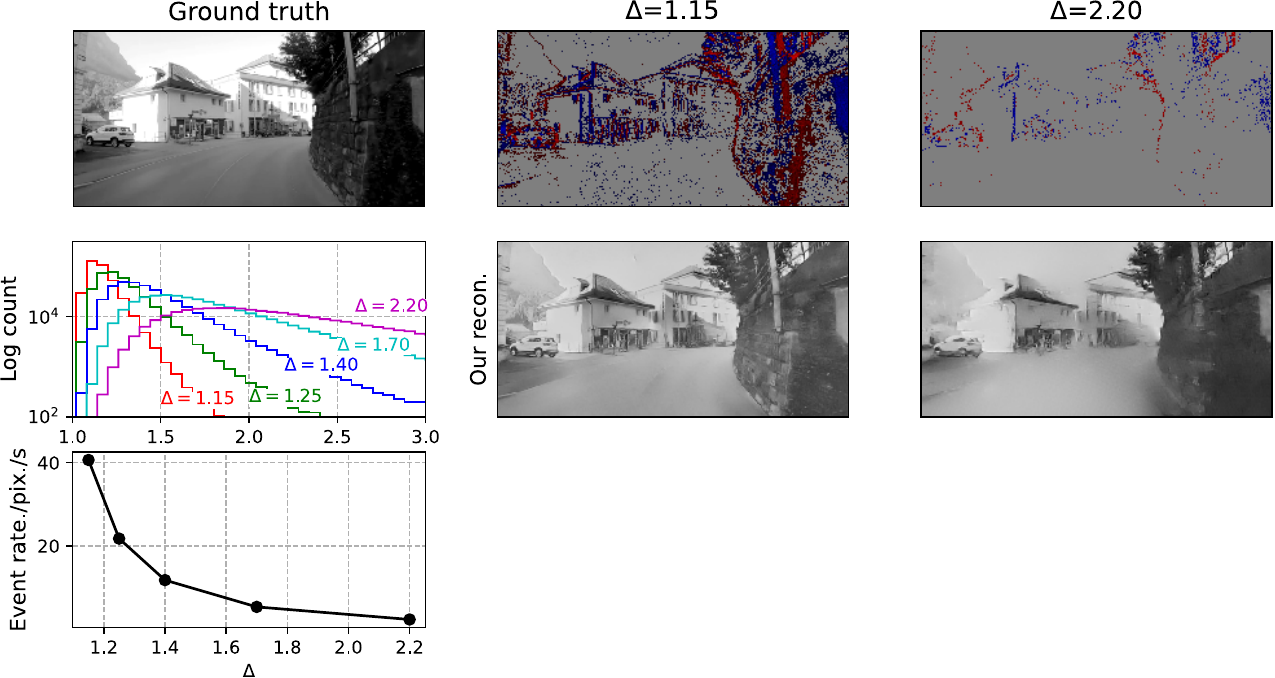}
\caption{
Exemplary ground truth image $\bfL$ and corresponding discretized events $\bfD^\Delta$ for different thresholds $\Delta$.
Reconstructions are given for the globally fixed threshold value $\Delta$. 
A histogram of \emph{effective} threshold values are given for 5 \emph{target} threshold values: 1.15, 1.25, 1.4, 1.7, 2.2.
Mean event rate decreases with target activation threshold increase. 
}
    \label{fig:different_deltas}
\end{figure}

\section{Methods}
\subsection{Event Simulator}
To generate training sequences we adopted the reconstruction framework from Rebecq~\emph{et al.}~\cite{rebecq2019high} and used the MS-COCO dataset~\cite{lin2014microsoft}.
We randomly draw 1'000 2D images, convert each to grayscale and scale them to simulate illuminance $\bfE$ from brightness $\bfL\in[0, 1]$ of $H\times W$ resolution: 
\begin{equation}
 \bfE = L_\text{rng}\bfL + L_\text{min}, ~~\text{where } L_\text{min}\sim U[50, 5'000], 
 ~ L_\text{rng}\sim U[100, 20'000].
\end{equation}

For each illuminance image $\bfE$ (in lux), a random but smooth-in-time sequence of affine transforms is applied via bicubic interpolation with symmetric extrapolation handling, yielding a frame sequence $\bfE_t$, ($t=1,\dots,T_h$).
To avoid the inverse crime scenario~\cite{KAIPIO2007493}, the event simulation is conducted at a higher frame rate than reconstruction, i.e. we produce the training sequence $\bfL_1, \dots, \bfL_{T_l}$ by picking every $q$=$10$-th frame of the original sequence of length $T_h$, where $T_h$=$qT_l$.

The sequence $\bfE_t$ can be considered a high frame-rate video input for the event simulator with the target activation threshold $\Delta$, producing the event stream:
\begin{equation}
    \{e_1^\Delta, \dots, e_K^\Delta\} \leftarrow \text{Simul}(\{\bfE_1, \dots, \bfE_{T_h}\}; \Delta),
    \label{eq:im2ev}
\end{equation}
where each event $e^\Delta_i=(\tau_i, x_i, y_i, s_i)$ is associated with a tuple consisting of event trigger time $\tau_i$, two spatial coordinates ($x_i\in\{1, \dots, W\}, ~ y_i\in\{1, \dots, H\}$) and a binary polarity of the intensity change ($s_i\in\{+1, -1\}$).
For example, the positive event is triggered when the contrast ratio exceeds the activation threshold: $E(\tau_i)/E(\tau_{i-1})\geq\Delta$.
We used a proprietary and internally-validated, physically accurate EVS simulator developed by Sony Research. 
This simulator is similar to V2E~\cite{hu2021v2e, Joubert2021EventCS}, as opposed to the more widely used ESIM~\cite{rebecq2018esim}, to leverage the more realistic modelling of the sensor's circuitry. 
Optics and non-idealities are also taken into account. 
The latter includes mismatch, electronic and background activity noise sources, frequency responses of the various circuit blocks in the pixel pipeline, refractory period and temperature dependence.

The event stream is then discretized into \emph{fixed-temporal-width} bins, producing the normalized event array sequence 
$
    \{\bfD^\Delta_1, \dots, \bfD^\Delta_{T_l}\} \leftarrow \mathcal{D}(\{e^\Delta_1, \dots, e^\Delta_K\}), 
$
where $\bfD^\Delta_i$, $\bfL_i$ have matching spatial dimensions $H\times W$ and 
\begin{equation}
    \bfD^\Delta_t(x, y) = \Delta\cdot\sum_{i: x=x_i, y=y_i} s_i \max\{1 - |t-\tau_i|, 0\},
\end{equation}
where the temporal frame length of the $\bfL_t$ sequence is assumed to be equal to 1 in terms of event timestamps.
I.e. we simply do a signed event count on a spatiotemporal grid, normalizing it with the activation threshold value $\Delta$ to roughly account for the contrast variation magnitude.
We similarly define the per-pixel event rate array in the following way:
\begin{equation}
    \bfC^\Delta_t(x, y) = \sum_{i: x=x_i, y=y_i} |s_i| \max\{1 - |t-\tau_i|, 0\}.
\end{equation}
It is worth noting that the target activation threshold in Eq.~(\ref{eq:im2ev}) defines the \emph{average} value of $\Delta$, while the effective per-pixel thresholds are random due to various circuit non-idealities~\cite{lichtsteiner2008a128x128}.
The effective threshold distribution, average event rates and exemplary discretized event frames can be seen in Fig.~\ref{fig:different_deltas}. 

\subsection{Design Considerations for Integrated Circuits}
Since we aim to optimize per-column distributions of the activation thresholds $\Delta$, in this section we explain how such control is implemented in our simulator and cover its feasibility from the chip-design perspective.

Biases refer to adjustable currents that determine key characteristics of the EVS pixel, such as the positive and negative thresholds $\Delta$, the refractory period, bandwidth, and others. 
In sensors like the DAVIS~\cite{brandli2014a240x180}, the bias currents for each pixel are produced outside of the pixel array using a bias generator~\cite{yang2012addressable} and are then mirrored to each pixel. 
Typically, bias lines are shared throughout the entire array, leading to uniform parameters for all pixels. Variations in these parameters are mainly attributed to mismatch and voltage drops in the wiring.

For the purpose of acquisition optimization we need to have spatial control of  bias parameters, however addressing individual pixels is impossible for large arrays due to physical wiring constraints. 
Therefore, we favor per-column control over per-row control, as it better suits the horizontal motion in car-mounted cameras.
We address individual lines of pixels through separate columns of bias lines which can be independently digitally switched  across a discrete set of current values by using a current-based Digital to Analog Converter (DAC) at the edge of the corresponding column. 
The bits deciding the output of the current-based DAC of each column can be obtained by an encoding coming from the sensor control network.
The digital switching of bias currents, including their settling time, is modelled in our simulator. 
At each change in bias value, the pixel state is flushed in order to prevent events from being generated by decreasing thresholds and residual charge. 
Therefore we additionally constrain our control network to update column states at 7.5\,Hz (every fourth step in the discretization) to mitigate the pixel state flushing effect contribution in mathematical modelling.
A detailed circuit implementation can be found in~\cite{patent_sony}.

\subsection{Image Reconstruction}
Using a simulated, paired training set of events $\bfD_t$ and ground truth brightness images $\bfL_t$, one can solve the reconstruction problem as a regression from events into brightness images using a recurrent convolutional architecture $\mathcal{V}_\theta$~\cite{scheerlinck2020fast}:
\begin{equation}
    \min_\theta \sum_{t\leq T_l}\mathcal{L}(\bfL_t, \mathcal{V}_\theta(\bfD_t)),
\end{equation}
where $\mathcal{L}$ is the perceptual image dissimilarity metric, namely LPIPS-VGG~\cite{zhang2018unreasonable}.

Now, assuming that the event simulation and subsequent discretization are provided for different values of $\Delta$, we have $N_c$ sequences of discretized events:
$\bfD_t^{\Delta_j}$, where $t=1,\dots,T_l$ and $j=1,\dots,N_c$, 
where $N_c$ is the number of considered distinct activation thresholds.
Then we assume that (for now data-independent) control over activation thresholds is provided by a one-hot sequence of binary masks $\bfM_t^j$ of shape $1\times W$, where the one-hot dimension is indexed by $j$.
Reconstruction learning for such predefined control is formulated as
\begin{equation}
    \min_\theta \sum_{t\leq T_l}\mathcal{L}(\bfL_t, \mathcal{V}_\theta(\tilde{\bfD}_t, \bfM_t)), ~~~
    \text{where } \tilde{\bfD}_t=\sum_{j\leq N_c}\bfD_t^{\Delta_j}\cdot\bfM_t^j.
    \label{eq:stochastic_recon}
\end{equation}
Here the fused event tensor $\tilde{\bfD}_t$ is constructed from precomputed events according to the control mask $\bfM_t^j$ and models the event stream that the reconstruction network would receive during real acquisition.

\subsection{On-The-Fly Control}
\begin{figure}[t]
    \centering
    \includegraphics[width=0.8\linewidth]{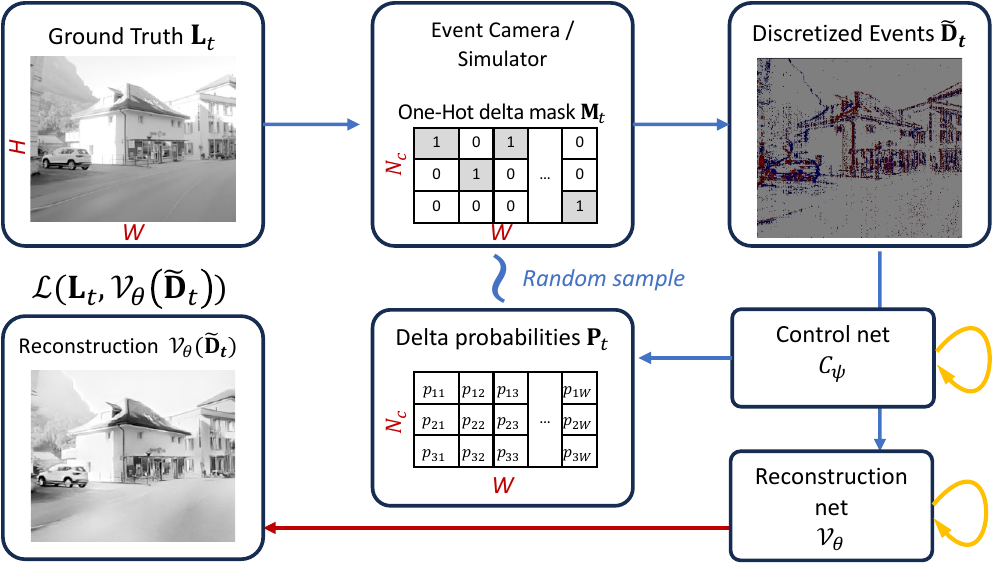}
    \caption{Illustration of the simulation-reconstruction-control feedback loop during training.
    In this example $N_c$=3.
    Recurrent states are indicated with yellow arrows.
    }
    \label{fig:event_mod}
\end{figure}

We introduce activation threshold optimization by building upon the randomized control formulation~(\ref{eq:stochastic_recon}), where the control masks $\bfM_t^j$ can be sampled randomly and unconditionally.
We propose to sample these mask entries from the \emph{categorical distribution} $\text{Cat}[\bfP_t]$ with probabilities $\bfP_t$ provided by the control network $\mathcal{C}_\psi$, which provides a feedback control over the sensor (cf. Fig.~\ref{fig:event_mod}):
\begin{align}
\begin{split}
    & \min_{\theta, \psi} \;\;
    \Exp{\bfM_t \sim \text{Cat}[\bfP_t]} \sum_{t\leq T_l}\;\;\mathcal{L}(\bfL_t, \mathcal{V}_\theta(\tilde{\bfD}_t, \bfM_t)) + \lambda_t\lambda_\text{max} \mathcal{B}(\tilde{\bfC}_t), \\
    &\text{where}\;\;\; \tilde{\bfD}_t=\sum_{j\leq N_c}\bfD_t^{\Delta_j}\cdot\bfM_t^j, \;\; 
      \tilde{\bfC}_t=\sum_{j\leq N_c}\bfC_t^{\Delta_j}\cdot\bfM_t^j,\\
    &\quad\quad\quad \bfM_t \sim \text{Cat}[\bfP_t],\;\; \bfP_t = \mathcal{C}_\psi(\tilde{\bfD}_{t-1}, \bfM_{t-1}, \tilde{\bfC}_{t-1}; \lambda_t),\\
    &\quad\quad\quad \lambda_t\sim U[0, 1],
\end{split}
\label{eq:big_opt}
\end{align}
where the event rate loss $\mathcal{B}$ is the mean event count of the fused sequence normalized by the event count corresponding to the lowest threshold $\Delta_1$:
\begin{equation}
    \mathcal{B}(\tilde{\bfC}_t)=\sum_{x,y}\tilde{\bfC}_t(x,y) \Big/ \sum_{x,y,i}\bfC^{\Delta_1}_i(x,y).
\end{equation}
Here the optimal balancing of the image reconstruction loss $\mathcal{L}$ and event rate loss $\mathcal{B}$ during training is provided by random values of $\lambda$, which can be set externally by the user during the inference stage to achieve the desired trade-off between performance accuracy and event bandwidth.
The image reconstruction loss in Eqs.~(\ref{eq:stochastic_recon}) and (\ref{eq:big_opt}) are identical, while the control network receives the fused event ($\tilde{\bfD}_{t-1}$) and event rate ($\tilde{\bfC}_{t-1}$) tensors from the previous step together with the penalty value $\lambda$ to be aware of the desired average event rate.
Visual illustration of the training loop setup is provided in Fig.~\ref{fig:event_mod}.

Note that both the reconstruction network $\mathcal{V}_\theta$ and control network $\mathcal{C}_\psi$ are abstracted from the complete event information contained in $\bfD^{\Delta_j}_t$ and $\bfC^{\Delta_j}_t$, as it would not be available during the inference.

\subsection{Optimization}
We approach the optimization of~(\ref{eq:big_opt}) in an alternating fashion by splitting gradients steps w.r.t. reconstruction net weights $\theta$ and control net weights $\psi$. 
Gradient calculation w.r.t. $\theta$ is straightforward, as we do not need to backpropagate through the discrete variable $\bfM_t$.
However, optimization w.r.t. $\psi$ is more challenging, as the influence of $\mathcal{C}_\psi$ on the cost function is via the expected discrete one-hot mask $\bfM_t$.

Similarly to reinforcement learning practices, note that the control optimization in~(\ref{eq:big_opt}) is formulated in the stochastic manner: i.e. we minimize the \emph{expected value} of the reconstruction cost over the sampling strategy defined by the activation threshold probabilities $\bfP_t$.
This allows us to see the problem~(\ref{eq:big_opt}) as a probabilistic relaxation of the challenging highly-dimensional discrete optimization problem, thus enabling stochastic gradient descent techniques.
Namely, we employ the Gumbel-Softmax trick~\cite{jang2017categorical, maddison2017the} to backpropagate through discrete masks $\bfM_t$. 
In short, this technique substitutes one-hot samples $\bfM_t^j$ with corresponding relaxed and differentiable samples from a Gumbel-Softmax distribution.
As a downside, the fused tensors $\tilde{\bfC}_t$ and $\tilde{\bfD}_t$ contain information leaked from all the corresponding $\bfC_t^{\Delta_j}$ and $\bfD_t^{\Delta_j}$.
However, we determined that this is not a practical concern, as the Gumbel-Softmax trick is used solely for calculating gradients with respect to $\psi$.

\subsection{Network Architectures and Training}
Generally, we use a 2D recurrent convolutional architecture for the reconstruction network $\mathcal{V}_\theta$ and a 1D recurrent convolutional network (since activation threshold control is per-column) for the control network $\mathcal{C}_\psi$.
Although different choices of network design yield different trade-offs in network efficiency, number of weights and reconstruction performance analyzed in detail in~\cite{scheerlinck2020fast, rebecq2019high}, in this work we focus on the conceptual improvement due to the sensor control and note that the control network can be hot-swapped to work with any other reconstruction network that is trained to predict image frames from the discretized events and one-hot $\Delta$ encoding mask.
We adapted the reconstruction network architecture from E2VID~\cite{rebecq2019high} to accommodate variable activation thresholds $\Delta$ and trained it using our simulator. 
Note that direct comparison with E2VID is impractical, as it was designed and trained for a globally constant $\Delta$. 

On each time step $t$ the reconstruction network $\mathcal{V}_\theta$ receives an array of size $(N_c$+$1)$$\times$ $N_x$$\times$$N_y$, where the channel dimension \emph{always} consists of a single layer corresponding to the event discretization $\tilde{\bfD}_t$ and $N_c$ layers consisting of a per-pixel one-hot encoding of the target activation thresholds used by the sensor.
The control network $\mathcal{C}_\psi$ receives a multichannel 1D input, where channels correspond to the per-column mean, max and standard deviation of $\tilde{\bfD}_t$, $|\tilde{\bfD}_t|$, $\tilde{\bfC}_t$ and a constant layer containing $\lambda$'s to provide the event rate conditioning.

In our experiments we used $N_c$=$5$ distinct target activation thresholds: $\Delta_i\in\{1.15, 1.25, 1.4, 1.7, 2.2\}$, with the fixed refractory period $t_\text{ref}=10^{-4}$\,s. 
We argue that increasing the amount of activation thresholds would result in diminishing returns, since the effective activation thresholds are randomly distributed around the target $\Delta$ (see. Fig.~\ref{fig:different_deltas}).

We use the Adam~\cite{kingma2014adam} optimizer for both steps with a learning rate of $5\cdot10^{-4}$, halving it on iteration 40'000 and terminating at iteration 75'000.
The Gumbel-Softmax relaxation temperature for the control network optimization step is set to $\tau_\text{sm}$=$\exp(0.5)$.
We use a batch size of 7, each training sequence consists of $T_l$=$64$ discretized time frames with spatial dimensions $H$=$W$=$96$, obtained by randomly cropping pre-generated sequences of length $T_l^\text{train}$=$100$ and $H^\text{train}$=$180$, $W^\text{train}$=$240$.

\begin{figure}
    \includegraphics[width=\linewidth]{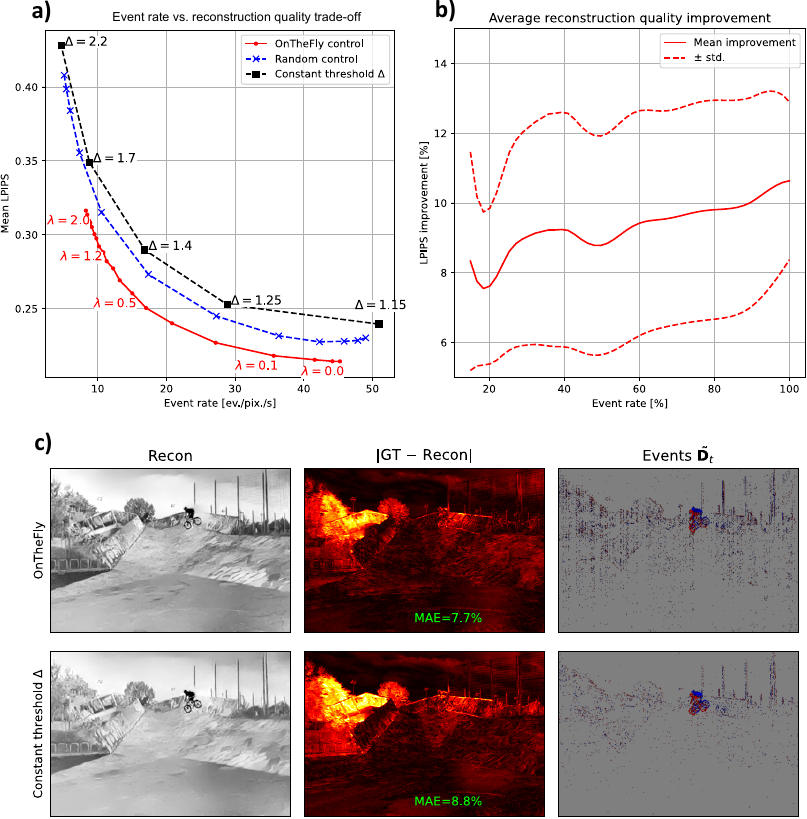}
\caption{
Reconstruction accuracy vs. event rate trade-off evaluation for the OnTheFly control.
a) LPIPS vs. event rate trade-off for the OnTheFly, Random and fixed globally constant threshold control regimes.
Random control is achieved by drawing the distribution of thresholds from the Dirichlet distribution and then sampling threshold configurations from it.
b) Mean image reconstruction quality improvement by OnTheFly control compared to the constant threshold configuration as a function of the event rate averaged over validation data. 
To compare different sequences, we assume the smallest value of $\Delta$ to achieve 100\% event rate.
The corresponding LPIPS values are then interpolated via cubic Hermite polynomials.
c) Exemplary reconstructed images together with corresponding discretized events for the constant activation threshold ($\Delta$=1.7) configuration and OnTheFly control ($\lambda$=1.8) with matching \emph{average} event rates.
}
    \label{fig:lpips_vs_event_rate}
\end{figure}

\section{Experiments}
In this section we present quantitative and qualitative evaluation of the proposed OnTheFly control for the event camera.
We use ten scenes captured at 60\,FPS containing a car driving as well as natural panoramas. 
The videos were cropped and downsampled to resolutions ranging from 300$\times$440 to 600$\times$300 pixels.

Fig.~\ref{fig:lpips_vs_event_rate}(a) illustrates the relationship between reconstruction quality and event rate across various acquisition control strategies. 
With a globally constant activation threshold, a decrease in the threshold value $\Delta$ consistently enhances image reconstruction accuracy (LPIPS) while also increasing the event rate.
The random control strategy, which dynamically but randomly alters threshold values, enables modulation of the event rate. 
However, this approach offers only a slight improvement in reconstruction accuracy.
Finally, the OnTheFly control strategy significantly enhances reconstruction quality and allows balancing between the event rate and reconstruction accuracy via the penalty parameter $\lambda$.
An exemplary reconstruction comparison for this sequence is shown in Fig.~\ref{fig:lpips_vs_event_rate}(c).
The range of the regularization parameter $\lambda$ used in this experiment does not cover the entire spectrum of possible event rates. 
Specifically, in the unpenalized scenario where $\lambda$=$0$, the control network does not generate the maximum achievable event rate, which would correspond to the smallest $\Delta$. 
This suggests that the lowest threshold values may encode an excess of information. Additionally, achieving lower event rates occasionally requires setting $\lambda>1$: a value beyond the training range of $[0, 1]$. 
Despite this, the control network appears capable of effectively extrapolating to these values without issue.
For the ``fully regularized'' case ($\lambda$=$2$) the lowest possible event rate (corresponding to $\Delta$=$2.2$) is not achieved because in the Lagrangian formulation of (\ref{eq:big_opt}) the event rate penalty would need to approach $+\infty$.

In Fig.~\ref{fig:lpips_vs_event_rate}(b) we show the quantitative improvement of the OnTheFly control strategy against constant fixed activation thresholds averaged over the validation set, confirming 8-10\% improvement in terms of the LPIPS metric.
We also establish a baseline for reconstruction accuracy using constant thresholds and then quantify the reduction in event rate achievable with the OnTheFly method to attain equivalent image accuracy.
Table~\ref{tab:evr_improvement}a illustrates that the maximum reduction, reaching 49\%, occurs at the lowest activation threshold. 
This reduction progressively diminishes to 4\% at higher threshold levels, where the scope for event dismissal is limited due to the already elevated threshold setting.
{Table~\ref{tab:evr_improvement}b shows that the control net improves the performance of a reconstruction net which was trained on a random control strategy: i.e. the reconstruction net is hot-swappable.}

\begin{table}[t]
\caption{a) Event rate reduction achieved by our On-The-Fly control, averaged over the validation set.
The average LPIPS reconstruction quality with the global constant activation threshold $\Delta$ is matched against the On-The-Fly reconstruction accuracy for which the event rate reduction is reported. 
b) the control net trained jointly with a downstream reconstruction net (tandem) is compared against using the control net with a reconstruction net trained on a random activation threshold distribution (hotswap). 
Event-rate regularization $\lambda$ was chosen to match the mean event-rate with global $\Delta$=$1.4$
}
\scriptsize
\vspace{-2mm}
(a)\hspace{5.5cm}(b)

\begin{tabular}{|p{2cm}|p{2cm}|}
\hline
\centering\textbf{Reference Recon Quality} & \centering\textbf{OnTheFly Event Rate Reduction} \tabularnewline \hline
$\Delta$=1.15& \qquad 49$\pm$ 8\%  \\ \hline
$\Delta$=1.25 &\qquad 28 $\pm$ 7\% \\ \hline
$\Delta$=1.4  &\qquad 17 $\pm$ 8\% \\ \hline
$\Delta$=1.7  &\qquad 4 $\pm$ 1\%  \\ \hline
$\Delta$=2.2  &\quad\qquad $\varnothing$   \\ \hline
\end{tabular}
\hfill
    \begin{tabular}{|c|c|c|c|c|}
        \hline
        \multicolumn{2}{|c|}{} & \multicolumn{3}{c|}{Dynamic Range} \\
        \cline{3-5}
        \multicolumn{2}{|c|}{} & x10 & x25 & x50 \\
        \hline
        \multirow{3}{*}{\shortstack{OnTheFly w. FT \\ (tandem training)}} 
        & SSIM & 0.57 & \textbf{0.70} & \textbf{0.80} \\
        & LPIPS & 0.39 & \textbf{0.30} & \textbf{0.25} \\
        & PSNR & 14.34 & \textbf{16.8} & \textbf{18.6}\\
        \hline
        \multirow{3}{*}{\shortstack{OnTheFfly wo FT \\ (hotswap rnd. trained net)}} 
        & SSIM & \textbf{0.58} & \textbf{0.70} & 0.79 \\
        & LPIPS & \textbf{0.38} & 0.31 & 0.27 \\
        & PSNR & 14.21 & 15.7 & 18.2 \\
        \hline
        \multirow{3}{*}{Globally constant $\Delta$=$1.4$} 
        & SSIM & 0.51 & 0.66 & 0.74 \\
        & LPIPS & 0.44 & 0.34 & 0.29 \\
        & PSNR & \textbf{14.37} & 15.4 & 15.8 \\
        \hline
    \end{tabular}
\label{tab:evr_improvement}
\end{table}

\begin{figure}[t]
    \includegraphics[width=\linewidth]{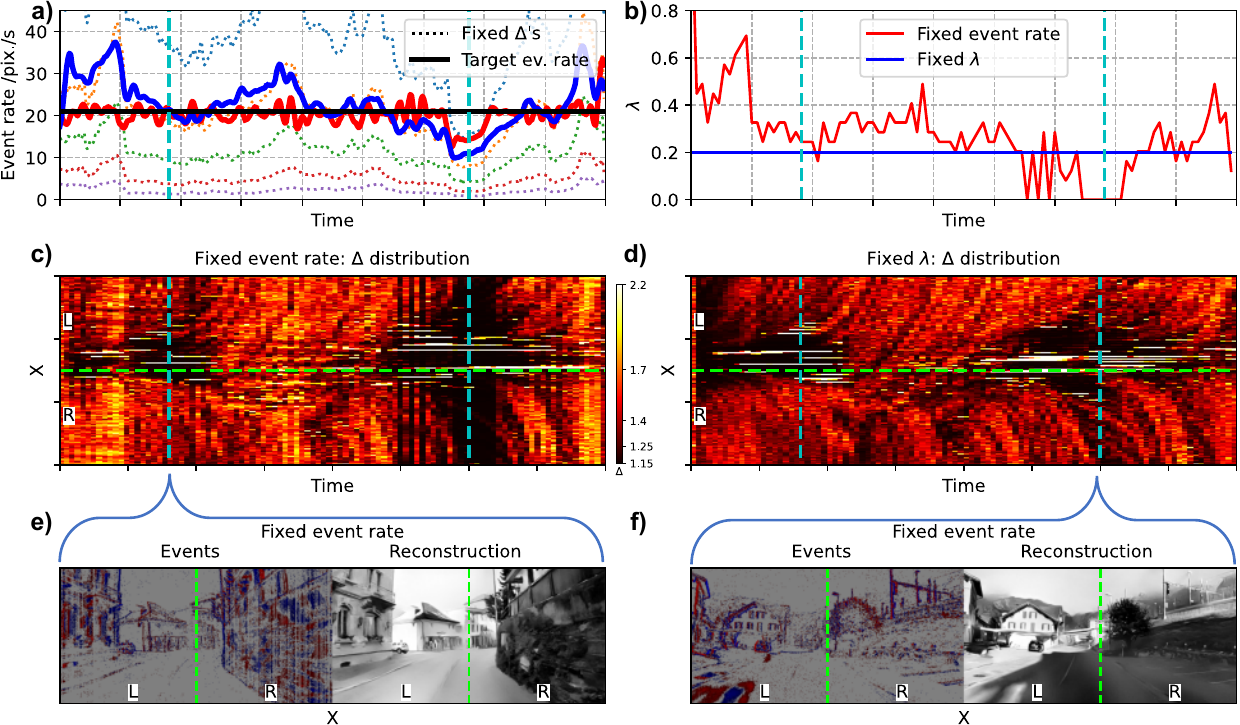}
    \vspace{-6mm}
\caption{
Fixed target event rate control.
a) Mean event rate for the control strategies, fixed global activation thresholds $\Delta$ and the target event rate.
Note that the fixed event rate control (red) is controlled not to exceed the target event rate.
b) Corresponding $\lambda$ values for fixed event rate and fixed $\lambda$ control.
c,d) Activation threshold $\Delta$ distribution per column over time for two OnTheFly control strategies.
e,f) discretized event frames and corresponding reconstructions for the fixed event rate control at time points indicated with dashed cyan lines.
Green dashed line indicates the middle column of the imaging sensor in $X$-Time and $X$-$Y$ panels. 
}
\vspace{-5mm}
    \label{fig:target_event_rate}
\end{figure}

\vspace{-3mm}
\subsection{Fixed Target Event Rate Control}
\vspace{-3mm}
Note that the event rate constraints in our training optimization problem~(\ref{eq:big_opt}) are imposed via a Lagrange penalty.
Such a formulation allows for a more straightforward approach for the optimization, avoiding costly and gradient-vanishing logits projections.
However, this approach does not allow for maintaining a fixed target event rate, as functional dependence between the value of $\lambda$ and the camera event rate is not known \emph{a priori}. 
To overcome this, we propose to choose the regularization value $\lambda^*_t$ dynamically, based on the running online estimate of the event rate per-threshold $\gamma^{\Delta_j}$ to maintain (on average) the target event rate value $\alpha$ as described in Algorithm~\ref{alg:runavg}.

\begin{algorithm}[t]
\caption{Target event rate control}
\label{alg:runavg}
\small
Input: $\alpha$~--- target event rate

$\gamma_0^{\Delta_j}\leftarrow 0$

For $t=1,\dots,T$:

\qquad$\gamma_{t}^{\Delta_j}\leftarrow\gamma^{\Delta_j}_{t-1} \cdot 0.2 + \frac{\sum_{x,y}\bfC_t^{\Delta_j}(x,y)\bfM_t^{j}(x,y)}{\sum_{x,y}\bfM_t^{j}(x,y)}\cdot 0.8$

\qquad$\lambda_{t+1}^*=\argmin_\lambda \left|\alpha - \sum_{j\leq N_c}\bfP_{t+1}^j\gamma_{t}^{j} \right|, $

$\text{\qquad\qquad~~ where~}\bfP_{t+1} = \mathcal{C}_\psi(\tilde{\bfD}_{t}, \bfM_{t-1}, \lambda).$

\text{\qquad Use }$\lambda_{t+1}^*$ for control net $\mathcal{C}_\psi$ at time $t+1$
\end{algorithm}

Results of such control are presented in Fig.~\ref{fig:target_event_rate}.
Note that the event rate of the fixed-event-rate control in general follows the target event rate, avoiding spikes due to camera motion as opposed to the fixed-$\lambda$ control.
The regularization values are adjusted to the scene dynamically: we see higher values of $\lambda$ when the scene is more dynamic (more events) and lower $\lambda$ for static segments (fewer events); this behavior is also followed in the activation thresholds $\Delta$ distribution.

Interestingly, Fig.~\ref{fig:target_event_rate} also allows us to intuitively interpret the optimal control produced by the OnTheFly network: as can be seen in Fig.~\ref{fig:target_event_rate}(c, d), lower threshold values are observed around the middle column of the sensor, i.e. the center of the field of view (FOV).
This scene is produced by a car-mounted camera, therefore the FOV center coincides with the vanishing-point of the optical system, and apparent motion as seen by the camera is smaller around it. Therefore, it would make sense to set these sensor pixel columns to be more sensitive.
We also observe that at the frame shown in Fig.~\ref{fig:target_event_rate}(e), the right part of the FOV contains a close-standing object (a wall), with high apparent motion, which is reflected in activation threshold asymmetry: $\Delta$'s on the right are larger than on the left in Fig.~\ref{fig:target_event_rate}(c,d).
Additionally, we can see curved ``arms'' of activation threshold groups produced by tracking edges and car turns.
Additionally, we see that the control network tends to produce a checkerboard pattern in the spatiotemporal $\Delta$ encoding to break neighbouring pixel coherence.

\vspace{-3mm}
\section{Conclusions and Discussion}
\vspace{-3mm}
We have presented a signal processing approach to the problem of event vision sensor control.
{Our proposed OnTheFly architecture allows for improving image reconstruction accuracy, indicating an increase in the amount of information encoded in the produced events.
The method of spatiotemporally-varying activation thresholds in event cameras necessitates further research regarding applications of standard vision algorithms to such novel data, e.g. by retraining the downstream network, or normalizing events by repeating those events with a large threshold $\Delta$ to match the nominal $\pm1$ activation.
}

The main limitation of our evaluation is the synthetic simulation of the EVS device, however we believe that this work can both motivate and accelerate the implementation of dynamically-controlled event cameras in silicon.
In this work we considered the fundamental EVS parameter i.e. the activation threshold. 
Future work can incorporate control of other parameters such as the refractory period and bandwidth or consideration of hybrid or color~\cite{taverni2018front} event sensors. 

Moreover, in future work more sophisticated spatial control schemes can be explored, such as joint column and row bias control.


%
%
\bibliographystyle{splncs04}
\bibliography{main}
\end{document}